\def\year{2017}\relax
\documentclass[letterpaper]{article}
\usepackage{aaai17}
\usepackage{times}
\usepackage{helvet}
\usepackage{courier}
\usepackage{tikz}
\usepackage{multirow}
\usepackage{subfigure}
\usepackage{amsfonts}
\usepackage{amsmath}
\usepackage{upref}
\begin{document}
%
\title{Deep Style Match for Complementary Recommendation}
\author{
Kui Zhao$^\dagger$, Xia Hu$^\ddagger$, Jiajun Bu$^\dagger$, Can Wang$^\dagger$\\
$\dagger$College of Computer Science, Zhejiang University\\
Hangzhou, China\\
\{zhaokui, bjj, wcan\}@zju.edu.cn\\
$\ddagger$Hangzhou Science \& Technology Information Research Institute\\
Hangzhou, China\\
hx@hznet.com.cn
}

\maketitle

\begin{abstract}
Humans develop a common sense of style compatibility between items based on their attributes. We seek to automatically answer questions like ``Does this shirt go well with that pair of jeans?'' In order to answer these kinds of questions, we attempt to model human sense of style compatibility in this paper. The basic assumption of our approach is that most of the important attributes for a product in an online store are included in its title description. Therefore it is feasible to learn style compatibility from these descriptions. We design a Siamese Convolutional Neural Network architecture and feed it with title pairs of items, which are either compatible or incompatible. Those pairs will be mapped from the original space of symbolic words into some embedded style space. Our approach takes only words as the input with few preprocessing and there is no laborious and expensive feature engineering. 
\end{abstract}

\section{Introduction}
\begin{figure}[t!]
\centering
\includegraphics[height=3.5cm]{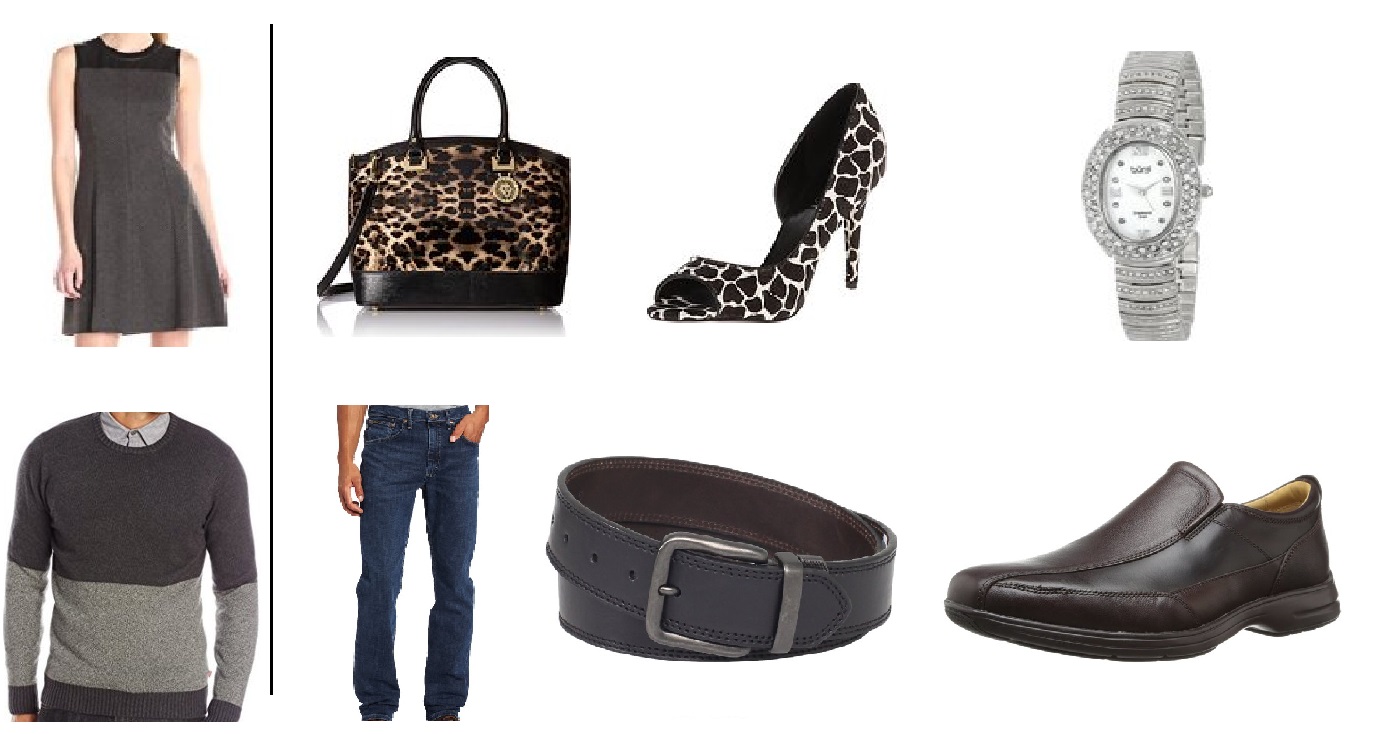}
\caption{An example for clothing outfits: on the left are query items and
on the right are corresponding complementary items.\label{fig:outfit}}
\end{figure}

We have a common sense of style compatibility between items
and can naturally answer questions like ``Does this shirt go well with that pair of jeans?''
These kind of style compatibility information can be exploited in many commercial applications, 
such as recommending items to users based on what they have already bought; or generating the whole 
purchase outfits (see Figure \ref{fig:outfit} for an example of clothes) to users querying certain items, if sufficient compatibility 
relationships between items are provided. 

To identify these compatibility relationships, existing methods such as frequent itemset mining \cite{han2000mining} 
attempt to generate match items automatically by analyzing historical purchasing patterns. 
However, frequent itemset mining relies on historical purchasing records to find items frequently 
purchased together and new items will inevitably suffer from the ``cold start'' problem \cite{schein2002methods}.

\begin{figure}[ht!]
\centering
\includegraphics[height=3.5cm]{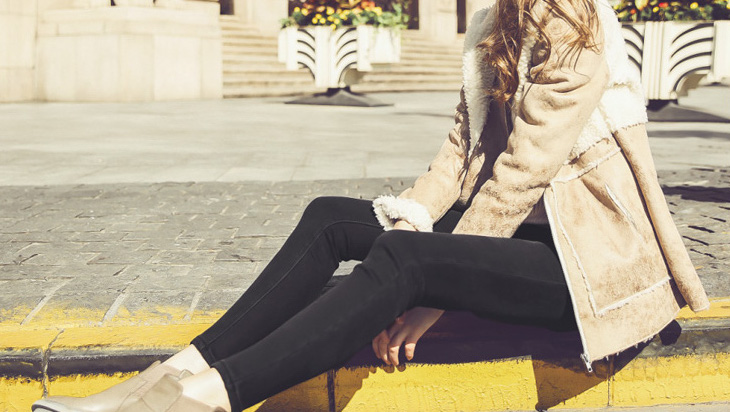}
\caption{A typical image from Taobao,
it actually is an image to show leggings.\label{fig:noise}}
\end{figure}

Recently, McAuley et al. \cite{mcauley2015image} and Veit et al. \cite{veit2015learning}
intend to discover the style match relationships between items using visual information presented in the images of items.
However, besides being computationally expensive, image-based matching methods 
are frequently plagued by the plentiful contents presented in images. 
For instance, Figure \ref{fig:noise} shows a part of the image for leggings from Taobao, 
which is the largest e-commerce platform in China. Besides leggings, 
the image also contains a coat, a pair of shoes and a very complex background etc. 
These contents will confuse the learning machines if only leggings are expected.

To overcome the limitations of existing methods,  
we propose in this paper a novel style match approach using the title descriptions of items in online stores. 
The basic assumption of our work is that online sellers will place most 
of the important attributes of a product in its title description, 
so that the product can be easily found by a keyword-based query. 
Therefore, the title description is a highly condensed collection of 
attribute descriptions for a product. 
So it is feasible to model compatibility between two items better if we are capable of mapping the title
pairs from the original space of symbolic words into some embedded style space.

We here design a Siamese Convolutional Neural Network architecture 
for matching title sentences. It will map two title sentences from an item pair into low-dimensional vectors respectively in parallel, 
which are then used to learn the compatibility between these two items in the style space. 
When designing the amalgamation part of Siamese CNN for computing compatibility, 
we have considered the ability of our model to be extended to  big data scenarios, 
which is critical for real-world recommendation applications. 
We test our approach on two large datasets: a Chinese dataset from Taobao provided by 
Alibaba Group and an English dataset from Amazon provided by \cite{mcauley2015image}. 
Our approach demonstrates strong performance on 
both datasets, which indicates its ability of learning human sense of style compatibility between items. 

\section{Related Work}
Finding complementary items has been studied for a long time. 
The early works can be traced back to frequent itemset mining \cite{han2000mining}, 
which generates match items automatically by analyzing history purchasing patterns. 
Frequent itemsets such as ``beer and diaper'' sometimes have nothing to do with compatibility. What's more, they are challenged by the ``cold-start'' problem, which means new products 
with no historical records are invisible to the algorithm \cite{schein2002methods}. 

Many approaches such as content-based recommendation or social recommendation are proposed to address this 
problem (see \cite{pazzani2007content} for a survey). 
Closely related to our work are \cite{mcauley2015image} and \cite{veit2015learning}, in which McAuley et al. and Veit et al. attempt to learn clothing style similarity based on their appearance in images. However, our work differs from \cite{mcauley2015image} and \cite{veit2015learning} in the following two aspects:
(1) we use title descriptions, which contain rich attribute information instead of item images; (2) the objective of our method is to find matching items from their attribute description instead of learning visual similarities from item images. 

\section{Problem formulation}
We here describe the problem in a formal way: 
given a query item set $Q=\{q_1, \cdots, q_m\}$ and a candidate item set $C=\{c_{1}, \cdots, c_{n}\}$, 
where each query item $q_i\in Q$ comes together with the compatibility 
judgements $\{y_{i_1}, \cdots, y_{i_n}\}$. 
The complementary item $c_j\in C$ is labeled with $y_{i_j}=1$ and $y_{i_j}=0$ otherwise. 
Our goal is to build a model to compute the compatibility probability between $q_i$ and $c_{j}$: 
\begin{equation}
\label{eq:pf}
P(y=1|q_i, c_{j}) = f(\phi(q_i,\theta_1), \phi(c_{j},\theta_1), \theta_2),
\end{equation}
where function $\phi(\cdot)$ is the sentence model mapping a title 
sentence into a low-dimensional representation vector 
and function $f(\cdot)$ computes the compatibility probability between two items in the style space. 
The parameter vectors $\theta_1$ and $\theta_2$ are learned in the training process. 

\section{Style Match}
The main building block of our approach is a sentence model based on CNN. 
This sentence model will map two title sentences from an item pair into low-dimensional vectors respectively in parallel, 
which are then used to learn the compatibility between two items in the style space.
\subsection{Sentence model}
\begin{figure}[ht!]
\centering
\includegraphics[height=5cm]{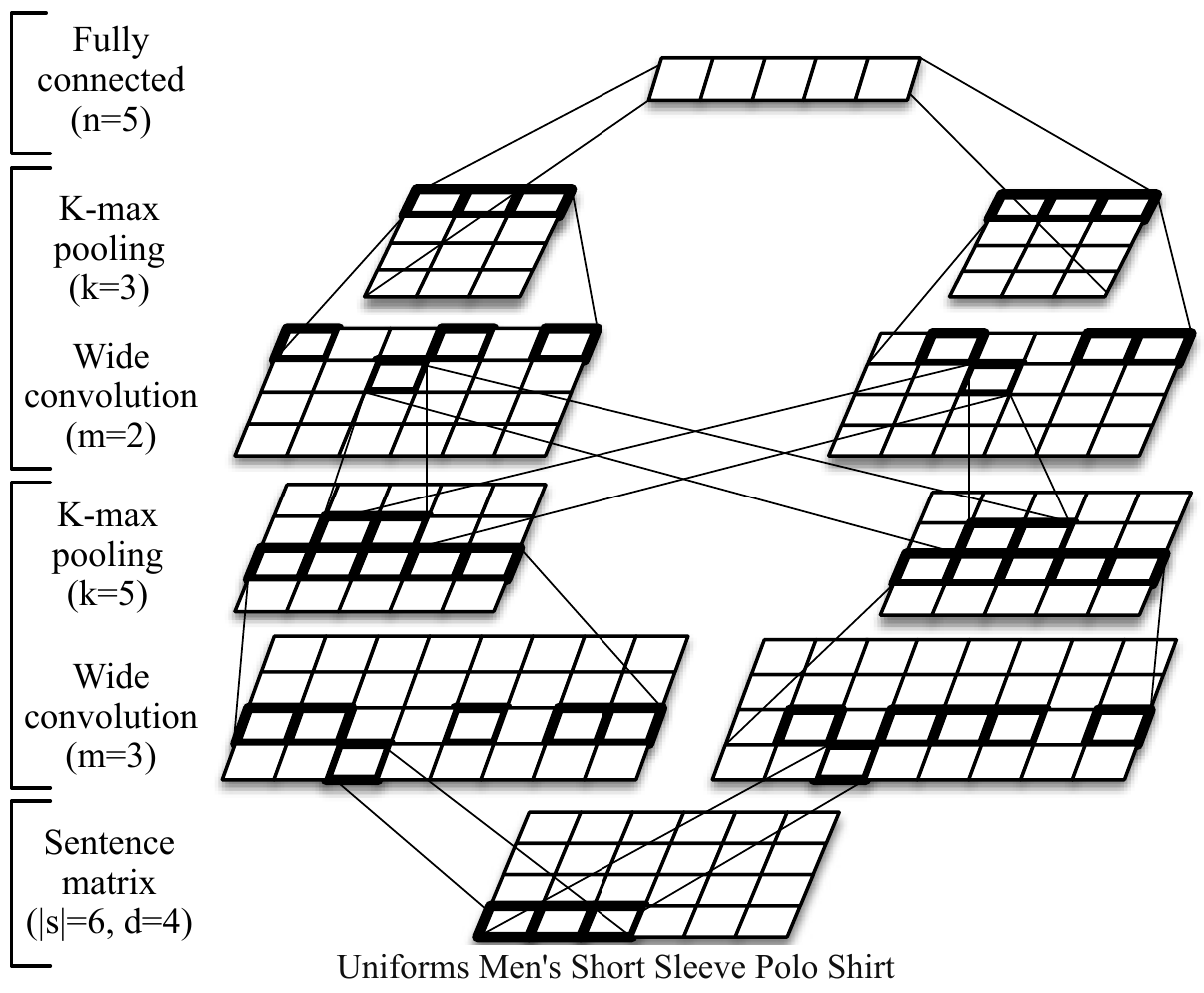}
\caption{The architecture of our sentence model.\label{fig:sentence_model}}
\end{figure}

We model sentences with function $\phi(\cdot)$, 
which is a convolutional architecture as shown in Figure \ref{fig:sentence_model}. 

In the following, we give a brief explanation of the main components in our Convolutional Neural Network. 
\subsubsection{Sentence matrix.}
Our sentence model takes a sentence ${\bf s}$ as the input, 
where it is treated as a sequence of raw words: $[s_1,\cdots,s_{|{\bf s}|}]$ 
and each word $s_i$ is from a vocabulary $V$. 

Firstly, each word $s_i$ is represented by a distributional representation vector 
${\bf w}_i \in \mathbb{R}^d$, looked up from the word-level embedding matrix ${\bf W} \in \mathbb{R}^{d\times |V|}$. 
Then a sentence matrix ${\bf S}\in \mathbb{R}^{d\times |{\bf s}|}$ can be built for the input sentence ${\bf s}$:
\begin{equation}
{\bf S}=
\begin{bmatrix}
\mid  & \mid  & \mid \\ 
 {\bf w}_1& \cdots & {\bf w}_{|{\bf s}|}\\ 
\mid & \mid & \mid
\end{bmatrix},
\end{equation}
where the $i$-th column is the distributional representation vector ${\bf w}_i$ 
for the $i$-th word in ${\bf s}$. 
The values in the embedding matrix ${\bf W}$ are parameters initialized with an unsupervised neural language model \cite{mikolov2013distributed} and sequentially optimized during training. 
The embedding dimension $d$ is a hyper-parameter of the model. 

As shown in Figure \ref{fig:sentence_model}, 
for the sentence ${\bf s}$=[Uniforms, Men's, Short, Sleeve, Polo, Shirt], 
when the embedding dimension $d$ is 4, the sentence matrix is a matrix in $\mathbb{R}^{4\times 6}$.

\subsubsection{Convolutional feature maps.}
Convolution can be seen as a special kind of linear operation and aimed to extract local patterns. 
We use the {\it one-dimensional convolution} to recognize discriminative word sequences from the input sentences. 
The one-dimensional convolution is an operation between 
two vectors ${\bf f}\in \mathbb{R}^{m}$ and ${\bf s}\in \mathbb{R}^{|{\bf s}|}$. 
The vector ${\bf f}$ is called as a {\it filter} of size $m$ and the vector ${\bf s}$  
is a {\it sequence} of size $|{\bf s}|$. The specific operation is to take the dot product of 
the vector ${\bf f}$ with each $m$-gram sliding along the sequence ${\bf s}$ and obtain 
a new sequence ${\bf c}$ where:
\begin{equation}
{\bf c}_j={\bf f}^{\rm T}{\bf s}_{j-m+1:j}. 
\end{equation} 
In practice, we usually add a bias $b$ to the dot product result:
\begin{equation}
\label{eq:con_with_bias}
{\bf c}_j={\bf f}^{\rm T}{\bf s}_{j-m+1:j}+b. 
\end{equation} 

There are two types of convolution depending on the allowed range of index $j$: {\it narrow} and {\it wide}. 
The narrow type restricts $j$ in the range $[m, |{\bf s}|]$ and 
the wide type restricts $j$ in the range $[1, |{\bf s}|+m-1]$. 
The benefits of wide type over the narrow type in text processing
are discussed in detail in \cite{blunsom2014convolutional}. 
Briefly speaking, unlike the narrow convolution 
where words close to margins are seen fewer times, 
wide convolution gives equal attention to all words in the sentence 
and so is better at handling words at margins. 
More importantly, a wide convolution always produces a valid 
non-empty result ${\bf c}$ even when $|{\bf s}|<m$. 
For these reasons, we use wide convolution in our model. 

The sentence matrix ${\bf S}$ is not just a sequence of single values but a sequence of vectors, 
where the dimension of each vector is $d$. 
So when we apply the one-dimensional convolution on the sentence matrix ${\bf S}$, 
we need a filter bank ${\bf F} \in \mathbb{R}^{d\times m}$ consisting of $d$ filters of size $m$
and a bias bank ${\bf B} \in \mathbb{R}^{d}$ consisting of $d$ baises. 
Each row of ${\bf S}$ is convoluted with the corresponding row of ${\bf F}$ 
and then the corresponding row of ${\bf B}$ is added to the convolution result. 
After that, we obtain a matrix ${\bf C} \in \mathbb{R}^{d\times (|{\bf s}|+m-1)}$:
\begin{equation}
\text{conv}({\bf S}, {\bf F}, {\bf B}): \mathbb{R}^{d\times |{\bf s}|} \to \mathbb{R}^{d\times (|{\bf s}|+m-1)}.
\end{equation}
The values in filter bank ${\bf F}$ and bias bank ${\bf B}$ are parameters optimized during training.
The filter size $m$ is a hyper-parameter of the model. 

In Figure \ref{fig:sentence_model}, 
after applying a $4\times 3$ filter bank and a bias bank of size $4$ on the $4 \times 6$ sentence matrix, 
we obtain an intermediate matrix of size $4 \times 8$. 

\subsubsection{Activation function.}
To make the network capable of learning non-linear functions, 
a non-linear activation $\alpha (\cdot)$ need to be applied in an element-wise way 
to the output of the preceding layer 
and a matrix ${\bf A}\in \mathbb{R}^{d\times (|{\bf s}|+m-1)}$ is then obtained:
\begin{equation}
\alpha({\bf C}): \mathbb{R}^{d\times (|{\bf s}|+m-1)} \to \mathbb{R}^{d\times (|{\bf s}|+m-1)}.
\end{equation}

Popular choices of $\alpha (\cdot)$ include: {\it sigmod}, {\it tanh} and {\it relu} (rectified linear defined as $\max (0, x)$). 
In practice, our experimental results are not very sensitive to the choice of activation,
so we choose {\it relu} due to its simplicity and computing efficiency. 
In addition, we can see the bias $b$ in (\ref{eq:con_with_bias}) plays the role of setting 
an appropriate threshold for controlling units to be activated. 

\subsubsection{Pooling.}
Pooling layer will aggregate the information in the output of preceding layer. 
This operation aims to make the representation more robust and invariant to small translations in the input. 
More importantly, pooling helps to handle inputs with varying size, e.g. processing sentences with uncertain length. 

For a given vector  ${\bf a}\in \mathbb{R}^{|{\bf a}|}$, traditional pooling aggregates it into a single value:
\begin{equation}
\text{pooling}({\bf a}): \mathbb{R}^{|{\bf a}|} \to \mathbb{R}.
\end{equation} 

The way of aggregating the information defines two types of pooling operations: {\it average} and {\it max}. 
Max pooling is used more widely in practice. 
Recently, max pooling has been generalized to {\it k-max pooling} \cite {blunsom2014convolutional}, 
in which $k$ max values are selected from the vector ${\bf a}$ and arranged in their original order: 

\begin{equation}
\text{k-pooling}({\bf a}): \mathbb{R}^{|{\bf a}|} \to \mathbb{R}^{k}, 
\end{equation} 
where $k$ is a hyper-parameter of the model. 

When we apply k-max pooling on the matrix ${\bf A}$, 
each row of ${\bf A}$ is pooled respectively and we obtain a matrix ${\bf P} \in \mathbb{R}^{d\times k}$: 
\begin{equation}
\text{k-pooling}({\bf A}): \mathbb{R}^{d\times |{\bf a}|} \to \mathbb{R}^{d\times k}. 
\end{equation}

In Figure \ref{fig:sentence_model}, after applying k-max pooling (with $k=5$) on 
the intermediate matrix of size $4\times 8$, we obtain a new intermediate matrix of size $4 \times 5$.

\subsubsection{Multiple feature maps.}
After a group of above operations, we obtain the first order {\it representation} 
learning to recognize the specific $m$-grams in the input sentence. 
To obtain higher order representations, we can use a deeper network 
by repeating these operations. The higher order representations can 
capture patterns of the sentence in much longer range. 

Meanwhile, we can also extend network to learning multi-aspect representations. 
Let ${\bf P}^i$ denote the $i$-th order representation. 
We can compute $K_i$ representations ${\bf P}^i_1,\cdots,{\bf P}^i_{K_i}$ 
in parallel at the same $i$-th order. Each representation ${\bf P}^i_j$ 
is computed by two steps. 
First, we compute convolution on each representation ${\bf P}^{i-1}_k$ at the lower order $i-1$
with the distinct filter bank ${\bf F}^i_{j,k}$ and bias bank ${\bf B}^i_{j,k}$  and then 
sum up the results. 
Second, non-linear activation and k-max pooling are applied to the summation result: 
\begin{equation}
{\bf P}^i_j=\text{k-pooling}(\alpha(\sum\limits_{k=1}^{K_{i-1}}\text{conv}({\bf P}^{i-1}_k, {\bf F}^{i}_{j,k}, {\bf B}^{i}_{j,k}))).
\end{equation}

In Figure \ref{fig:sentence_model}, there are two representations at the first order and two 
representations at the second order: ${\bf P}^1_1\in \mathbb{R}^{4\times 5}, {\bf P}^1_2\in \mathbb{R}^{4\times 5}$ 
and ${\bf P}^2_1\in \mathbb{R}^{4\times 3}, {\bf P}^2_2\in \mathbb{R}^{4\times 3}$.  

\subsubsection{Full connection.}
Full connection is a linear operation 
to combine all representations at the highest order into a single vector. 
More specifically, for the highest order representations 
${\bf P}^h_1,\cdots,{\bf P}^h_{K_h}$ (assume ${\bf P}^h_k\in \mathbb{R}^{d\times l})$, 
we first flat them into a vector ${\bf p}\in\mathbb{R}^{K_h\times d\times l}$. 
Then we transform it with a dense matrix ${\bf H} \in \mathbb{R}^{(K_h\times d\times l)\times n}$:
\begin{equation}
\label{fc}
{\bf x}={\bf p}^{\rm T}{\bf H},
\end{equation}
where ${\bf x}\in \mathbb{R}^n$ is the final representation vector. 
The values in matrix ${\bf H}$ are parameters optimized during training. 
The representation size $n$ is a hyper-parameter of the model.

In Figure \ref{fig:sentence_model}, we finally represent the input sentence with 
a vector of size $n=5$. 


\subsection{Matching items}
\begin{figure}[ht!]
\centering
\includegraphics[height=2.7cm]{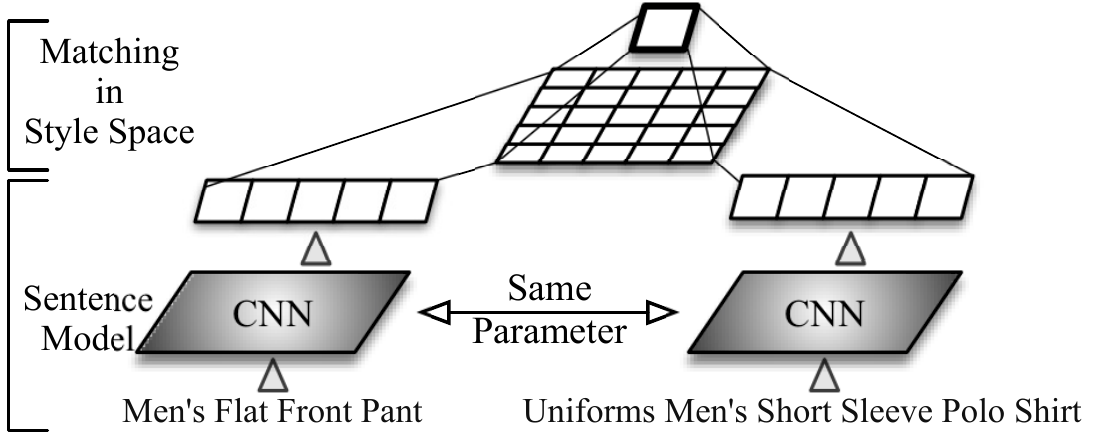}
\caption{The whole architecture of our model.\label{fig:style_match}}
\end{figure}

We compute the compatibility probability between two items 
with function $f(\cdot)$, which is a Siamese Convolutional Neural Network as shown in Figure \ref{fig:style_match}. 
 
Siamese setup is introduced by Hadsell et al. \cite{hadsell2006dimensionality} and used widely in learning distance metrics. 
When designing the amalgamation part of our model, 
we have considered its scalability for big data scenarios, 
which is critical for real-world applications. 

\subsubsection{Style space.} 
For two given items $q$ and $c$, after generating 
the representation vectors ${\bf x}_q\in \mathbb{R}^n$ and ${\bf x}_c\in \mathbb{R}^n$
of their title sentences respectively, we compute the compatibility probability between them as follow:
\begin{equation}
\label{eq:match}
\begin{aligned}
P(y=1|q,c)&=\sigma({\bf x}_q^{\rm T}{\bf M}{\bf x}_c+b)\\
&=\frac{1}{1+e^{-({\bf x}_q^{\rm T}{\bf M}{\bf x}_c+b)}},
\end{aligned}
\end{equation}
where ${\bf M}\in \mathbb{R}^{n\times n}$ is a matrix and $b$ is a scalar. 
We  call ${\bf M}$ as the {\it compatibility matrix} 
and the space spanned by ${\bf M}$ as the {\it style space}. 
After transformation ${\bf x}'_q={\bf x}_q^{\rm T}{\bf M}$, 
${\bf x}'_q$ represents the item which is most style compatible to $q$.  
We seek items whose representations are close to $\bf{x}'_q$ under linear kernel distance. 
The values in compatibility matrix ${\bf M}$ and the bias $b$ are parameters 
optimized during the training. 

On the other hand, ${\bf x}_q^{\rm T}{\bf M}{\bf x}_c$ in (\ref{eq:match}) can be 
viewed as a noisy-channel model, which has been widely used in the information retrieval and 
QA system \cite{echihabi2003noisy} \cite{bordes2014open}. 

\section{Recommendation}
In recommendation applications, 
we are usually given a query item set $Q=\{q_1, q_2,\cdots, q_m\}$ 
and a candidate item set $C=\{c_1, c_2,\cdots,c_n\}$, 
where the query item set is relatively small and the candidate item set is usually very large. 
For each query item $q_i$, we intend to query its $K$ most complementary items from 
the candidate set $C$ and rank them from high compatibility to low compatibility. 
When the item candidate set is very large, 
it is inefficient and even unacceptable to compute the compatibility for all item pairs $(q_i,c_j)$ and then sort them. 

Our approach can be easily extended to handle these big data scenarios.
Given two items $q$ and $c$, we first generate their representation vectors ${\bf x}_q$ and ${\bf x}_c$ respectively. 
Then their compatibility probability is computed according to (\ref{eq:match}). 
We notice that the function $\sigma (\cdot)$ in (\ref{eq:match}) is a monotonic increasing function and $b$ is a learned constant. 
Thus for a query item $q$ and two candidate items $c_1, c_2$, we have: 
\begin{equation}
\label{eq:ls}
\begin{aligned}
P(y=1|q,c_1)&\le P(y=1|q,c_2) \\ 
\Leftrightarrow {\bf x}_q^{\rm T}{\bf M}{\bf x}_{c_1}&\le {\bf x}_q^{\rm T}{\bf M}{\bf x}_{c_2}. 
\end{aligned}
\end{equation}

Based on this property, we transform the original problem of querying the $K$ most complementary items 
of $q$ from the item candidate set $C$ 
into another problem, namely searching {\it $K$ nearest neighbors} of ${\bf x}'$ (${\bf x}'_q={\bf x}_q^{\rm T}{\bf M}$) 
from $\{{\bf x}_{c_1}, \cdots, {\bf x}_{c_n}\}$ under the linear kernel distance. 
It is well known as Maximum Inner Product Search (MIPS). 
There are many methods solving MIPS efficiently on the large scale data, 
such as tree techniques \cite{ram2012maximum} and 
hashing techniques \cite{shrivastava2014asymmetric} \cite{shen2015learning} etc.  


\section{Training}
We train the model to maximize the likelihood of a observed relationship training set $\mathcal{R}$, 
where $r_{ij}\in \mathcal{R}$: 
\begin{equation}
r_{ij}=\begin{cases}
1 & \text{, if items {\it i} and {\it j} are compatible;} \\ 
0 & \text{, otherwise.}
\end{cases}
\end{equation} 
Maximizing the likelihood is equal to minimizing the binary-cross entropy loss function:
\begin{equation}
L=-\sum\limits_{r_{ij}\in\mathcal{R}}\left[r_{ij}\log(p)\\
+(1-r_{ij})\log(1-p)\right],
\end{equation}
where $p=P(y=1|i,j)$. 

The parameters to be optimized in our network are $\theta_1,\theta_2$, which have been mentioned above:
\begin{equation}
\theta_1=\{{\bf W}, {\bf F}, {\bf B}, {\bf H}\}\text{ and }
\theta_2=\{{\bf M}, b\},
\end{equation}
namely the word embeddings matrix ${\bf W}$, filter bank ${\bf F}$, bias bank ${\bf B}$, 
dense matrix ${\bf H}$, compatibility matrix ${\bf M}$ and compatibility bias $b$. 
Note that there are multiple filter banks and bias banks to be learned. 

In the following sections, we present several crucial details for training our deep learning model. 
\subsection{Regularization}
To alleviate the overfitting issue, we use a popular and efficient regularization 
technique named {\it dropout} \cite{srivastava2014dropout}. 
Dropout is applied to the flatted vector ${\bf p}$ (presented in ({\ref{fc})) before 
transforming it with the dense matrix ${\bf H}$. 
A portion of units in ${\bf p}$ are randomly dropped out by setting them to zero 
during the forward phase, which is helpful for preventing the feature co-adaptation. 
The dropout rate is a hyper-parameters of the model. 

\subsection{Hyper-parameters}
\label{hp}
The hyper-parameters in our deep learning model are set as follows: 
the embedding dimension is $d=100$; the size of filters at the first order representation is $m=3$; 
the number of max values selected by k-max pooling at the first order representation is $k=5$; 
the size of filters at the second order representation is $m=2$; 
the number of max values selected by k-max pooling at the second order representation is $k=3$; 
the dimension of the vector used to represent the sentence is $n=100$; 
the dropout rate is $p=0.2$. 
What's more, there are $K_1=100$ representations computed in parallel at the 
first order representation and  $K_2=100$ representations computed in parallel at the second order representation. 

\subsection{Optimization}
To optimize our network, we use the Stochastic Gradient Descent (SGD) algorithm with shuffled mini-batches. 
The parameters are updated through the back propagation framework with Adagrad rule \cite{duchi2011adaptive}. 
The batch size is set to 256 and the network is trained for 20 epochs. 
The training progress will be early stopped if there is no more update 
to the best loss on the validation set for the last 5 epochs. 

We train our network on a GPU for speeding up. 
A Python implementation using Keras\footnote{http://keras.io} 
powered by Theano \cite{Bastien-Theano-2012} 
can process 428k text pairs per minute on a single NVIDIA K2200 GPU. 

\section{Experiments}
We evaluate our method on two large datasets: 
a Chinese dataset from Taobao and an English dataset from Amazon. 
\subsection{Datasets}
\subsubsection{Taobao.} This dataset is collected from Taobao.com and provide 
by Alibaba Group\footnote{http://tianchi.aliyun.com/datalab/index.htm}. 
It includes a Clothing category and there are about 406k compatibility relationships covering 61k items. 
The compatibility relationships in this dataset are labelled 
manually by clothes collocation experts. 

\subsubsection{Amazon.} This dataset is collected from Amazon.com and provided by \cite{mcauley2015image}. 
Though it includes multiple categories, 
in order to investigate the performance of our approach on both datasets, 
we mainly focus on the Clothing category. 
In this category, there are about 12 million compatibility relationships covering 662k items. 
Unlike the Taobao dataset, the compatibility relationships in Amazon dataset are not labelled manually. They are the co-purchase data from Amazon's recommendations \cite{linden2003amazon}. 

\subsection{Setup}
Our goal is to differentiate compatibility relationships from non-compatibility ones. 
We consider all positive relationships (compatibility) and generate random non-relationship distractors 
of the equal size. That is to say the ratio between positive and negative samples in the dataset is 50:50. 
Then we separate the whole dataset into training, validation and testing sets according to the ratios 80:10:10. 
Although we do not expect overfitting to be a serious issue in our experiment with the large training set, we still carefully tune our model on the validation set to avoid overfitting on testing set. We compare our approach against baselines from two aspects: 
visual one and non-visual ones. 

\subsubsection{Visual baseline.} We take the method in \cite{veit2015learning} as the visual comparison 
since it is also in the end-to-end fashion. 
In particular, we consider the specific setting configured with GoogLeNet 
and naive sampling for two considerations. First, in all situations of their experiments, 
GoogLeNet \cite{szegedy2015going} outperforms AlexNet \cite{krizhevsky2012imagenet}. 
Second, naive sampling means sampling randomly from the dataset, 
which is consistent with the setup in our experiments. 
We experiment their method on the Taobao dataset and 
take the results on the Amazon dataset directly from \cite{veit2015learning}.

\subsubsection{Non-visual baselines.} We take three methods as the non-visual comparison:

1) {\it Naive Bayes on Bag of Words (NBBW)}. We treat the title sentences from an item pair as the bag-of-words and 
feed Naive Bayes classifier with it as the feature vector; 

2) {\it Random Forest on Bag of Words (RFBW)}. Random Forest is capable of modeling extremely complex classification surface. 
We apply Random Forest classifier on the bag-of-words representation of the title sentences from an item pair; 

3) {\it Random Forest on Topic Model (RFTM)}. 
For the given item pair $\{q, c\}$, we first generate the topic representations ${\bf x}_q, {\bf x}_c$ of items $q,c$ 
by LDA model \cite{blei2003latent} respectively, where the topic number is set as 100. 
Then we concatenate them into a single feature vector ${\bf x}_{q,c}$ and process Random Forest classifier on it.  

The implementation of Naive Bayes and Random Forest is taken from scikit-learn \cite{scikit-learn}. 
We turned their parameters to obtaine the best loss on the validation set. 

There is no preprocess on images and texts. All results reported in the following section is on the testing set. 

\subsection{Results}
\subsubsection{Comparison to baselines.}
Tabel \ref{tb:auc} shows the corresponding areas under the ROC curves of compatibility prediction on the testing set . 
The results show clearly that our approach outperforms all other baselines. 

\begin{table}[ht!]
\begin{center}
\begin{tabular}{c|c c c c c}
\hline
 Methods & &  Taobao & & Amazon  &  \\
\hline
Visual    & & 0.579  & & 0.770 &  \\
 NBBW   & &  0.712 & & 0.820 &  \\
RFBW   & &  0.807 & & 0.931  &  \\
RFTM   & &   0.796 & &  0.893 &  \\
Ours    & &  {\bf 0.891} & &  {\bf 0.983} &  \\
\hline
\end{tabular}
\end{center}
\caption{AUC scores for all methods.}
\label{tb:auc}
\end{table}
\begin{table*}[ht!]
\begin{center}
\begin{tabular}{c c | c c | c c c}
\hline
 Category & AUC   & Category & AUC & Category & AUC\\
\hline
Automotive    & 0.922 & Electronics   & 0.948 &  Patio Lawn \& Garden    & 0.966 \\
 Baby   & 0.917 &  Grocery \& Gourmet Food   &  0.959 & Pet Supplies    & 0.972 \\
Beauty   & 0.935  &  Health \& Personal Care    &  0.929 &  Sports \& Outdoors    & 0.912 \\
Books   &  0.897 & Home \& Kitchen    & 0.949  & Tools \& Home Improvement    & 0.952 \\
CDs \& Vinyl  &  0.815  & Movies \& TV  & 0.878   &  Toys \& Games    & 0.985 \\ 
Cell Phones \& Accessories & 0.969   & Musical Instruments    & 0.983 & Video Games    & 0.890  \\
 Digital Music   & 0.818 &   Office Products    & 0.974 & \\
\hline
\end{tabular}
\end{center}
\caption{AUC scores for compatibility prediction on twenty top-level categories from Amazon dataset.}
\label{tb:all_auc}
\end{table*}
The visual method collapses on the Taobao dataset because 
unlike Amazon, Taobao is a Consumer to Consumer (C2C) platform and 
has few strict requirement about quality of item images uploaded by users. 
A majority of images are like Figure \ref{fig:noise}, 
where the information is mixed up and confusing to learning machines. 
In contrast, the title description is a highly condensed collection of more attributes besides appearances 
with few noises. When using title descriptions, 
a simple method like Naive Bayes on Bag of Words can 
achieve an acceptable performance and 
a more sophisticated method like Random Forest on Bag of Words 
can generate competitive results. 
However, using topic models on title descriptions is not a good idea since most title descriptions are short texts. One important reason why our approach achieves better performance 
is that our approach can recognize specific $m$-grams and  
more complicated patterns not captured by bag-of-words models. 
For instance, the complementary styles of the item titled with 
``white shirt with blue stripes'' and ``blue shirt with white stripes'' 
are very different, but they have the same bag-of-words representation. 

The performance upper bound of our approach on Taobao dataset is 
limited by the segmentation quality of Chinese. This is one of the reasons that all AUC scores on 
the Taobao dataset are lower than that on the Amazon dataset. 

\subsubsection{Tuning sentence model.}
There are several crucial setups in the sentence model: 
1) the word embedding dimension $d$; 
2) the sentence representation dimension $n$;
3) whether the word embedding matrix ${\bf W}$ is initialized or not. 

\begin{figure}[ht!]
\centering
\includegraphics[height=4.5cm]{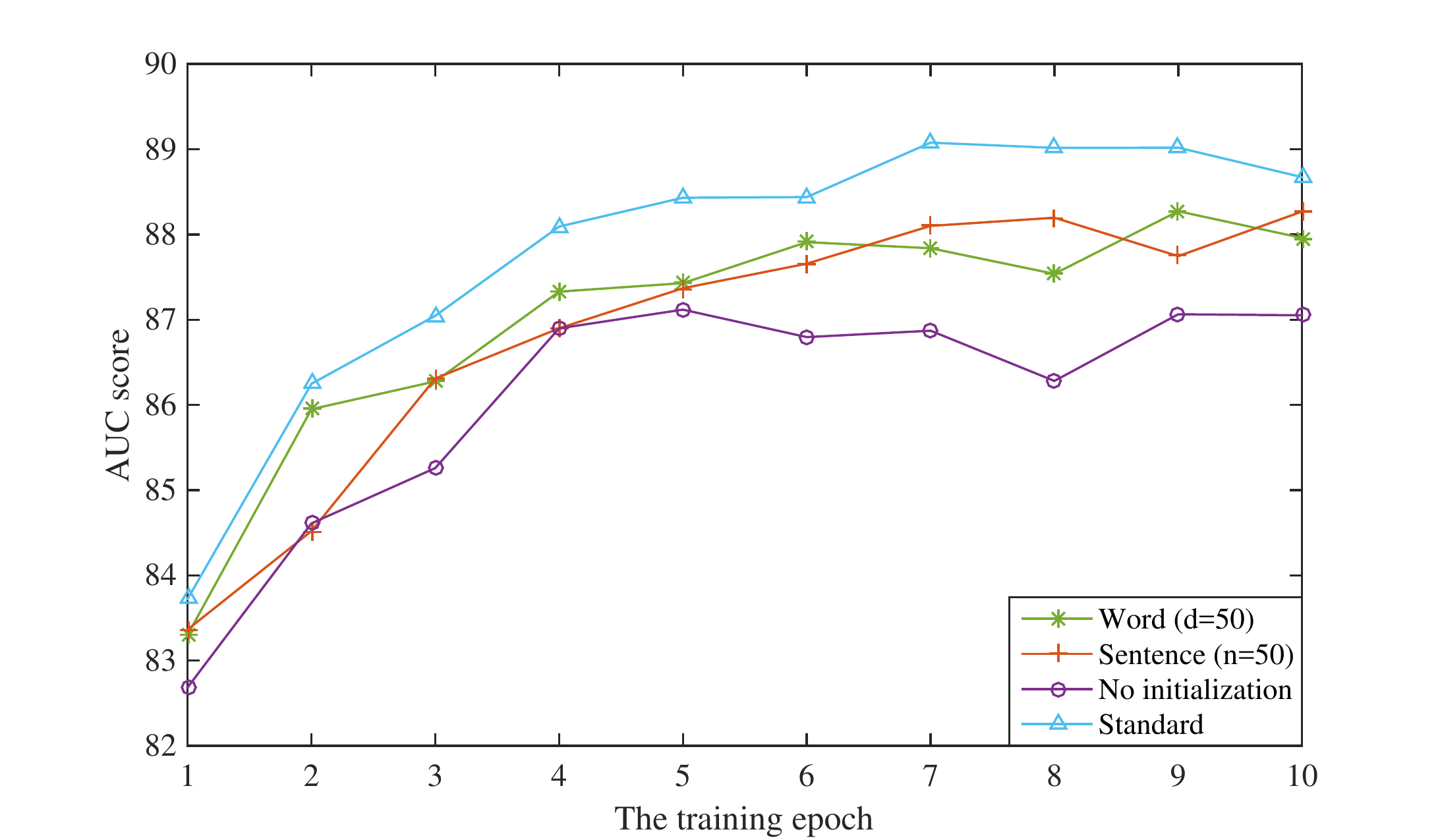}
\caption{Convergence processes of our approach with sentence models under different setups.\label{fig:diff}}
\end{figure}

We show the first ten epochs of training processes of our model on the Taobao dataset with sentence models under different setups in 
Figure \ref{fig:diff}, where the standard setup means that we set $d=100, n=100$ and 
initialize the word embedding matrix with an unsupervised neural language model \cite{mikolov2013distributed}. 
We can see that the setup with larger $d$ or $n$ claims better performance. 
Furthermore, the performance can get better when we continue to increase the value of $d$ or $n$. 
In practice, there is a tradeoff between the performance and resources requirement 
according to specific situations. 
What's more, initializing the word embedding matrix ${\bf W}$ with an unsupervised neural language model 
is indeed benefit to the convergence rate and the final performance. 

\subsection{Discussion}

\subsubsection{Toward general match.}
While the previous section mainly focuses on clothes matching, 
we also train classifiers on the other twenty top-level categories from the Amazon dataset 
and present the results in Table \ref{tb:all_auc}. 
As can be seen, we obtain good accuracy in predicting compatibility relationships in 
a variety of categories. 
What's more, we have also tried to train a single model to predict compatibility relationships for all categories. 
There appears to be no ``silver bullet'' and the result is dissatisfactory: the AUC score of that single model is only 0.694. 

The comparison across categories is particularly interesting. 
Our approach performs relatively poor on the categories ``CDs \& Vinyl'' 
and ``Digital Music'' since the content of music is too rich to be 
described very clearly in a short title description. 
In contrast, the title description is long enough to describe 
an item from the category `Musical Instruments' clearly and thus our approach performs very well on that. 
In a word, the better titles can describe the attributes of items in a category, 
the higher performance can be achieved on that category by our approach. 

\section{Conclusions}
In this paper, we present a novel approach to model the human 
sense of style compatibility between items. 
The basic assumption of our approach is that most of the important attributes for 
a product in an online store are included in its title description. 
We design a Siamese Convolutional Neural Network architecture to 
map the title descriptions of an item pair from the original space of symbolic words into some embedded style space. 
The compatibility probability between items can be then computed in the style space. 
Our approach takes only words as the input with few preprocessing and 
requires no laborious and expensive feature engineering. 
Moreover, it can be easily extended to big data scenarios with KNN searching techniques. 
The experiments on two large datasets confirm our assumption and 
show the possibility of modeling the human sense of style compatibility.  

There are several interesting problems to be investigated in our future work: 
(1) we would like to use more sophisticated sentence models 
without injuring the simplicity of our approach; 
(2) we are wondering whether it is possible to use the text and image information simultaneously, 
e.g. hybrid model or mapping the texts and images of items into the same embedded space 
for mutual retrieval and matching. 

\section{Acknowledgments}
We would like to thank Alibaba Group and Julian McAuley for providing the valuable datasets. 
This work is supported by Zhejiang Provincial Natural Science Foundation of China (Grant no. LZ13F020001), 
Zhejiang Provincial Soft Science Project (Grant no. 2015C25053), 
National Science Foundation of China (Grant nos. 61173185, 61173186).
\bibliographystyle{aaai}
\bibliography{aaai}  
\end{document}